\documentclass[conference]{IEEEtran}
\IEEEoverridecommandlockouts

\usepackage{cite}
\usepackage{amsmath,amssymb,amsfonts}
\usepackage{algorithmic}
\usepackage{graphicx}
\usepackage{textcomp}
\usepackage{subfigure}
\usepackage{subcaption} 
\usepackage{footmisc}
\usepackage{booktabs}
\usepackage{colortbl}
\usepackage{multirow}

\newenvironment{keywords}{\textbf{Keywords:}}{}
\def\BibTeX{{\rm B\kern-.05em{\sc i\kern-.025em b}\kern-.08em
    T\kern-.1667em\lower.7ex\hbox{E}\kern-.125emX}}
\begin{document}

\title{DySCo: Dynamic Semantic Compression for Effective Long-term Time Series Forecasting}

\author{
\IEEEauthorblockN{Xiang Ao, Yinyu Tan, Mengru Chen}
\IEEEauthorblockA{\textit{School of Software Engineering} \\
\textit{Beijing Jiaotong University}\\
Beijing, China \\
ao.xiang.axel@outlook.com, 24301001@bjtu.edu.cn, 24301100@bjtu.edu.cn}
}
\maketitle

\begin{abstract}
Time series forecasting (TSF) is critical across domains such as finance, meteorology, and energy. While extending the lookback window theoretically provides richer historical context, in practice, it often introduces irrelevant noise and computational redundancy, preventing models from effectively capturing complex long-term dependencies. To address these challenges, we propose a \textbf{Dynamic Semantic Compression (DySCo)} framework. Unlike traditional methods that rely on fixed heuristics, DySCo introduces an \textbf{Entropy-Guided Dynamic Sampling (EGDS)} mechanism to autonomously identify and retain high-entropy segments while compressing redundant trends. Furthermore, we incorporate a \textbf{Hierarchical Frequency-Enhanced Decomposition (HFED)} strategy to separate high-frequency anomalies from low-frequency patterns, ensuring that critical details are preserved during sparse sampling. Finally, a Cross-Scale Interaction Mixer(CSIM) is designed to dynamically fuse global contexts with local representations, replacing simple linear aggregation. Experimental results demonstrate that DySCo serves as a universal plug-and-play module, significantly enhancing the ability of mainstream models to capture long-term correlations with reduced computational cost.
\end{abstract}


\begin{figure}[htbp]
    \centering
    \includegraphics[width=1\linewidth]{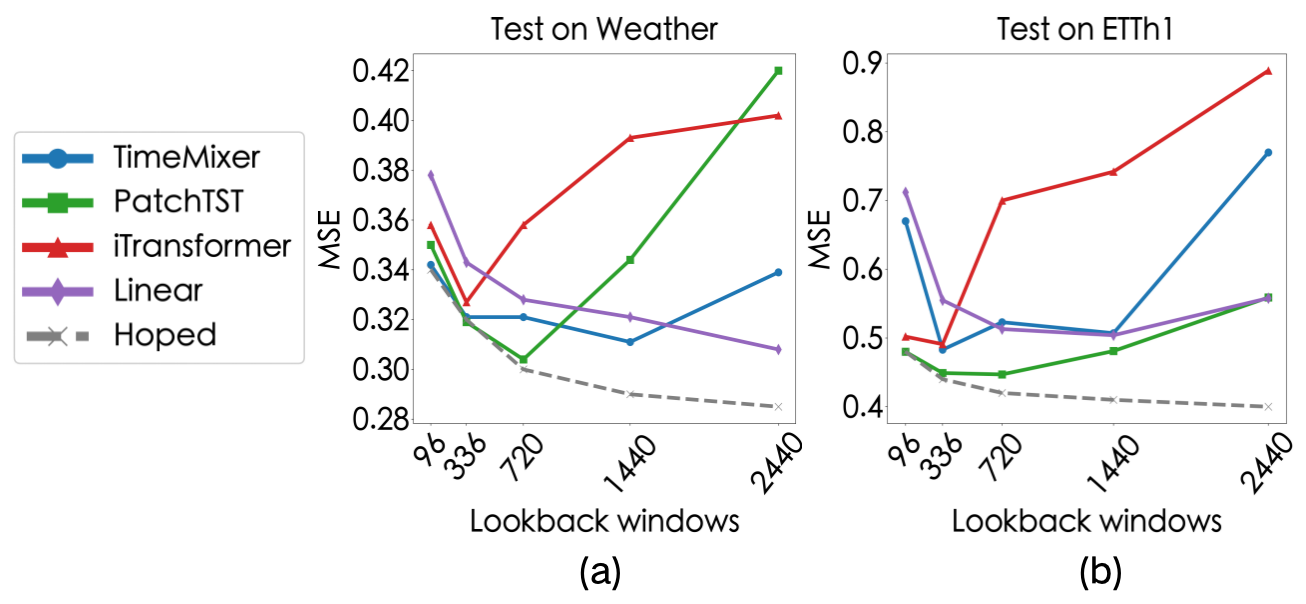}
    \caption{Predictive performance (measured by MSE) under different lookback windows. The "Hoped" denotes the results conforming to common sense.}
    \label{fig:1}
\end{figure}

\section{Introduction}
\label{sec:intro}

In industrial applications, predicting information over a future time period based on historical data is often essential to support operational decision-making. This demand is prevalent across domains such as finance, meteorology, transportation, and healthcare \cite{1,2,3,4}. These extensive practical requirements have continuously propelled the development of time series forecasting (TSF) technology.

Time series data are characterized by strict chronological ordering and inherent temporal regularities, such as periodicity, amplitude shifts, and evolving trends \cite{5}. Intuitively, extending the lookback window should provide richer context for prediction. However, empirical evidence suggests a paradox that simply increasing the input length often fails to improve, or even degrades, predictive accuracy (see Figure \ref{fig:1}). This is primarily due to the information redundancy and noise accumulation inherent in long historical sequences, where the correlation between distant past and current steps weakens significantly \cite{6}.

\begin{figure}[htbp]
    \centering
    \subfigure[Long lookback window]{%
        \label{fig:21}
        \includegraphics[width=0.45\textwidth]{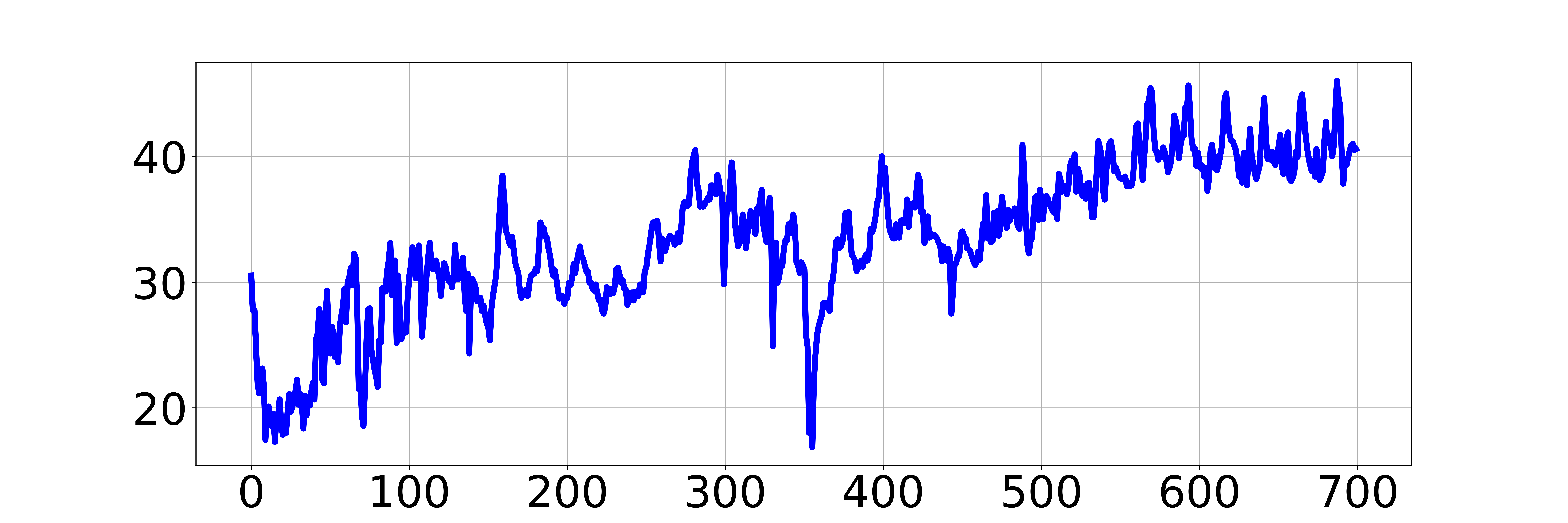}
    }
    \\
    \subfigure[Short lookback window]{%
        \label{fig:22}
        \includegraphics[width=0.45\textwidth]{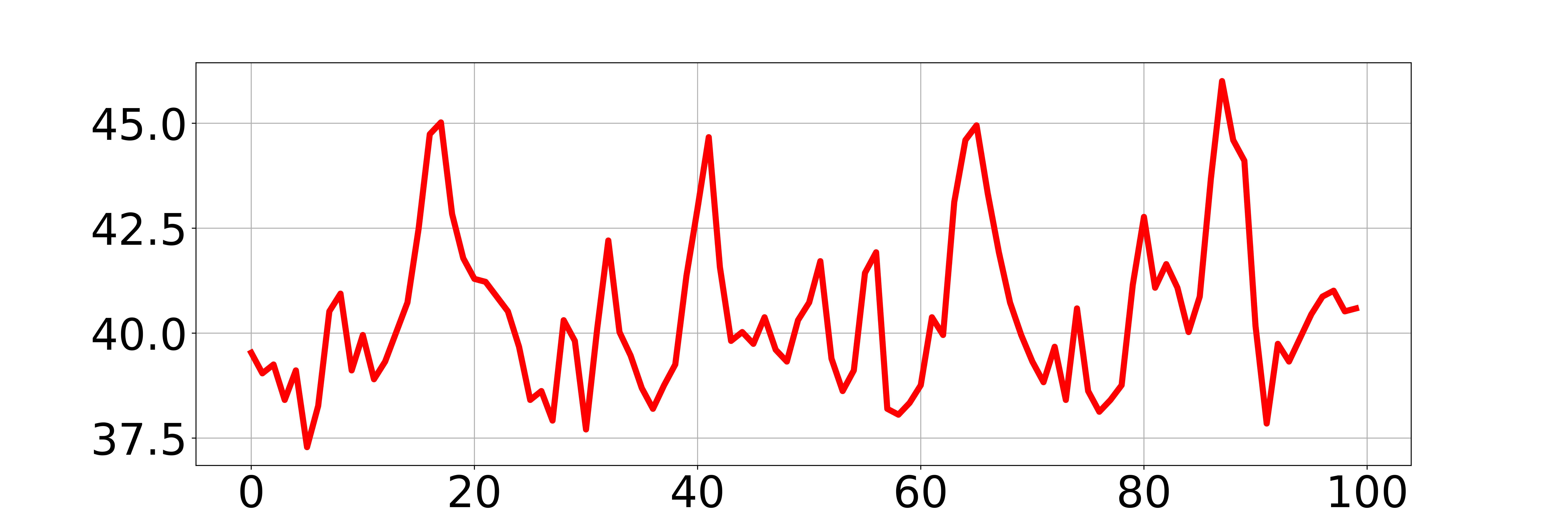}
    }
    \caption{Competition of different lookback windows. The short lookback window is derived from the long lookback window spanning steps 600 to 700. Note how the long window captures the trend but obscures the local periodicity visible in the short window.}
    \label{fig:2}
\end{figure}

Nevertheless, blindly truncating history is not a viable solution. As shown in Figure \ref{fig:2}, certain critical regularities are strictly long-term dependent. For instance, detecting weekly rush-hour patterns in urban traffic requires a receptive field covering at least one full week, utilizing only recent short-term data makes capturing such periodicity impossible. Therefore, the core challenge lies in effectively distinguishing valuable signals from irrelevant noise within long sequences.

To tackle these challenges and transcend the limitations of rule-based sampling, we propose the Dynamic Semantic Compression (DySCo) method. Unlike traditional methods that compress sequences based on distance, DySCo introduces a learnable paradigm to distill long sequences. Specifically, DySCo incorporates Entropy-Guided Dynamic Sampling (EGDS) equipped with an importance evaluation mechanism Hierarchical Frequency-Enhanced Decomposition (HFED) for multi-granularity modeling, and a Cross-Scale Interaction Mixer(CSIM) for adaptive fusion.

EGDS adaptively identifies and retains high-entropy segments (rich in information) while compressing redundant trends, ensuring that critical precursors, regardless of how far back they occur, are preserved. HFED then constructs a multi-granularity pyramid to separate high-frequency anomalies from low-frequency patterns. Finally, the CSIM replaces simple linear superposition with a context-aware gating mechanism. It dynamically weights and fuses global contexts with local representations, ensuring that the prediction benefits from both the stability of long-term trends and the sensitivity of short-term details.

To validate the effectiveness of DySCo, we integrated it into mainstream TSF models as a universal plug-and-play module. Experimental results demonstrate that DySCo significantly enhances the ability of base models to capture long-term correlations without incurring excessive computational costs.

The remainder of this paper is organized as follows. Section \ref{sec:related_work} briefly reviews related work. Section \ref{sec:method} details the proposed DySCo framework and its core components. Section \ref{sec:efficiency} provides a theoretical analysis regarding computational complexity and parameter efficiency. Section \ref{sec:exp} presents extensive experimental evaluations and visualizations. Finally, Section \ref{sec:conclusion} concludes the paper.

\begin{figure*}
    \centering
    \includegraphics[width=1\linewidth]{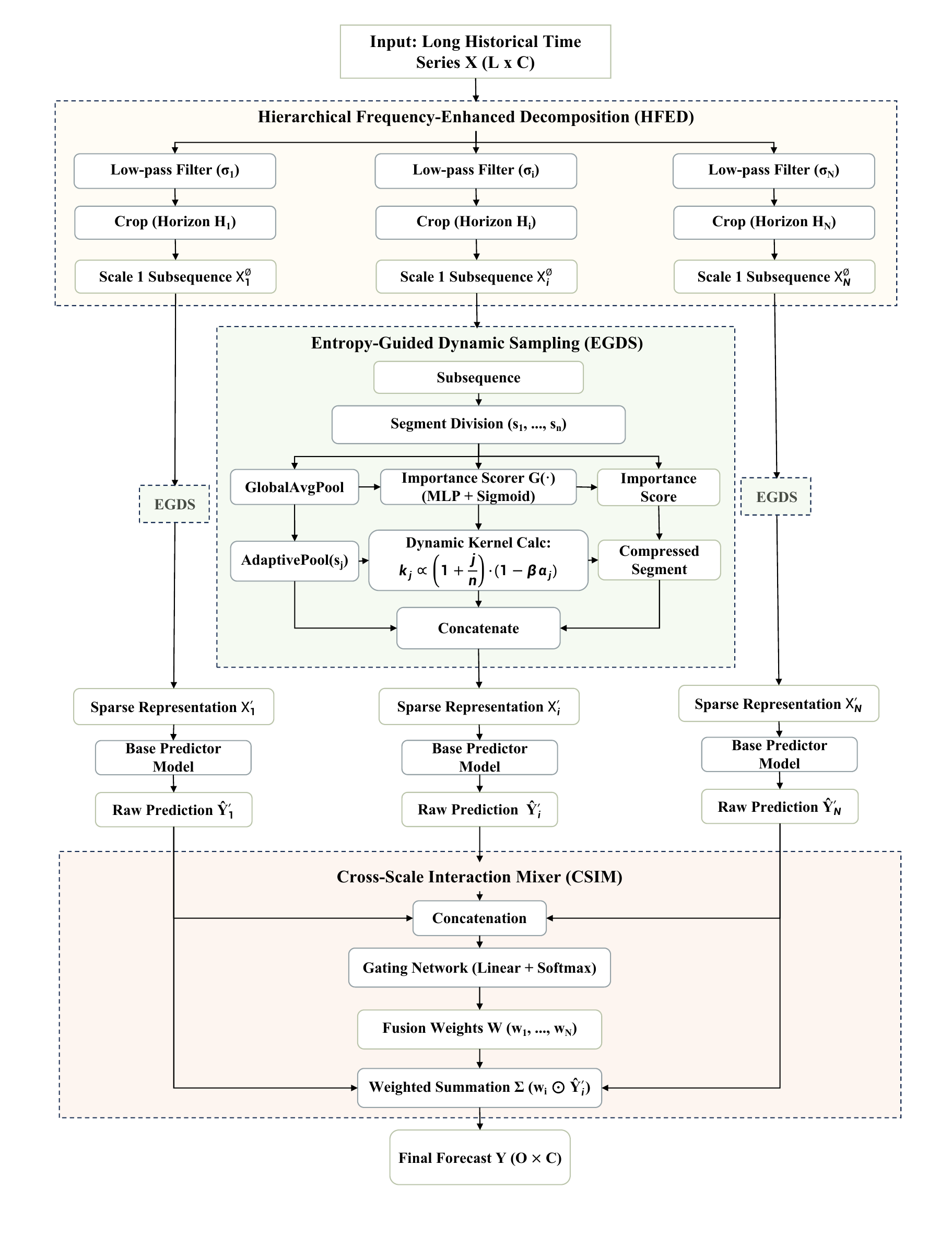}
    \caption{Framework of DySCo. The left panel illustrates the overall basic framework of DySCo, which comprises the HFED module and prediction-related components; the right panel details the workflow of the EGDS module.}
    \label{fig:model}
\end{figure*}


\section{Related Work}
\label{sec:related_work}
Existing TSF models employ divergent strategies to balance dependency capture and noise reduction. RNN-based models (e.g., LSTM, GRU) use gating to discard distant information \cite{7,8}, which mitigates noise but often compromises long-term dependencies due to gradient vanishing. Conversely, Linear and Transformer-based architectures (e.g., N-BEATS, PatchTST) treat historical points with uniform priority \cite{9,10,11,12}. Despite their efficacy, these methods suffer from high computational complexity and are prone to overfitting long-term noise. Recent state space models and foundation models attempt to optimize this efficiency-accuracy trade-off \cite{mamba_tsf2025}. Crucially, current sampling or compression techniques still rely on fixed heuristics, lacking the semantic adaptability to dynamically retain high-value information based on data context \cite{15,micn,timemixer++}.

\section{DySCo: Dynamic Semantic Compression}
\label{sec:method}

Time series forecasting aims to predict future sequence data \(Y \in \mathbb{R}^{O \times C}\) based on historical time series observations \(X \in \mathbb{R}^{L \times C}\), where $L$ is the length of the historical sequence, $O$ is the prediction horizon, and $C$ is the number of channels. To overcome the limitations of rigid lookback windows, we propose the DySCo framework. Unlike traditional rule-based methods, DySCo introduces a learnable paradigm to distill critical dependencies. 

The framework consists of three key components: 
1) \textbf{HFED}, which decomposes sequences into multi-granularity representations.
2) \textbf{EGDS}, which performs dynamic compression guided by learnable importance scores rather than fixed heuristics. 
3) \textbf{CSIM}, which dynamically aggregates predictions through a context-aware gating mechanism. The overall framework is illustrated in Figure \ref{fig:model}.

\subsection{Spectral-Temporal Dichotomy Analysis}
The motivation for DySCo stems from the inherent frequency-temporal trade-off in time series modeling\cite{frequency,xlinear}. Observing a time series from a long lookback window (Figure \ref{fig:21}) versus a short lookback window (Figure \ref{fig:22}) reveals a spectral dichotomy. 

Short-term local windows are dominated by high-frequency Components (e.g., periodic fluctuations and immediate noise), requiring dense sampling to preserve details. Conversely, long-term global windows primarily exhibit low-frequency components (e.g., trend evolution and seasonality, which possess high information redundancy).
Existing methods typically treat these components uniformly. Our key insight is that optimal sampling should be content-adaptive: retaining high resolution for high-entropy regions (unpredictable variations) while aggressively compressing low-entropy regions (predictable trends), regardless of their temporal distance. The rationale is that predictable patterns exhibit high redundancy and remain easily capturable even after compression, whereas complex, unpredictable variations necessitate dense preservation to guarantee information integrity.

\subsection{Hierarchical Frequency-Enhanced Decomposition}HFED is designed to construct a multi-granularity pyramid from the original sequence. For an input sequence $X \in \mathbb{R}^{L \times C}$, instead of simple truncation, we employ a hierarchical decomposition strategy to disentangle information across different scales.

Formally, we define a set of hierarchical lookback horizons $\mathcal{H} = \{H_1, H_2, \ldots, H_N\}$, where $H_1 < H_2 < \ldots < H_N = L$. Each $H_i$ represents a distinct receptive field, ranging from short-term local details to long-term global trends.

First, to adhere to the frequency-temporal trade-off, we assign a scale-specific bandwidth $\sigma_i$ to each layer. For longer horizons (larger $H_i$), a smaller $\sigma_i$ is applied to filter out high-frequency noise and for shorter horizons, a larger $\sigma_i$ is used to preserve sharp variations. The hierarchical representation $X_i^\phi$ is then generated by:

\begin{equation}
\begin{aligned}
\tilde{X}_i &= \mathcal{F}\text{low}(X, \sigma_i), \quad \tilde{X}_i^\phi &= \text{Crop}(\tilde{X}_i, H_i) = \tilde{X}_i[L-H_i+1 : L, :]
\end{aligned}
\end{equation}
where $\mathcal{F}{low}$ denotes the low-pass filter, and $\text{Crop}(\cdot, H_i)$ extracts the most recent $H_i$ time steps from the filtered sequence.

Specifically, to maintain computational efficiency while achieving strict frequency separation, we implement $\mathcal{F}_{low}$ as a time-domain moving average filter (i.e., 1D Average Pooling). The scale-specific bandwidth $\sigma_i$ inversely dictates the kernel size $k_i$ of the filter. For a given channel, the filtered output at time step $t$ is mathematically formulated as:
\begin{equation}
\tilde{X}{i, t} = \frac{1}{k_i} \sum{j=0}^{k_i-1} X_{t-j}
\end{equation}
where a smaller bandwidth $\sigma_i$ (applied to larger lookback horizons) corresponds to a larger kernel size $k_i$. This aggressively smooths out high-frequency local fluctuations, yielding a clean low-frequency global trend. 

Conversely, for the shortest local horizon (e.g., $H_1$), a maximum bandwidth $\sigma_1$ is assigned, corresponding to a minimal kernel size (e.g., $k_1 = 8$). This design ensures that essential high-frequency information—such as sharp variations and localized anomalies—is maximally preserved relative to the heavily smoothed global trends, preventing the loss of critical short-term signals.

This design ensures that high-frequency information—such as sharp variations and localized anomalies—is fully preserved in the smallest scale $X_1^\phi$, preventing the loss of critical short-term signals.This mechanism ensures that the resulting pyramid $\mathbb{X}^\phi = \{X_1^\phi, \ldots, X_N^\phi\}$ captures clean global trends at larger scales and sharp local details at smaller scales, effectively preventing aliasing effects during subsequent sampling.

\subsection{Entropy-Guided Dynamic Sampling}
\label{sec:egds}

Traditional sparse sampling relies on linear interpolation, operating on the heuristic that ``older data is less important.\cite{m1,m2,6}'' We argue this assumption is suboptimal for periodic or non-stationary data, as critical signals may appear in the distant past. Therefore, we propose EGDS, which modulates the sampling density based on the information density score of sequence segments.

First, similar to the basic approach, the filtered sequence $X_i^\phi$ is divided into $n$ consecutive segments $\{s_1, \ldots, s_n\}$. To evaluate the semantic value of each segment, we introduce a learnable importance scorer, denoted as $\mathcal{G}(\cdot)$. The scoring logic proceeds in feature abstraction and probability mapping. For each segment $s_j$, we first obtain a compact feature summary via global pooling, which is then projected into a scalar probability $\alpha_j \in [0, 1]$ by $\mathcal{G}$:
\begin{equation}
\label{eq:score}
\alpha_j = \mathcal{G}(s_j) = \mathcal{S}(\text{MLP}(\text{GlobalAvgPool}(s_j)))
\end{equation}
where $\mathcal{S}(\cdot)$ denotes the Sigmoid activation function. It is worth noting that rather than calculating strict mathematical entropy, the MLP is designed to act as a proxy for information density, effectively serving as a learnable importance scorer. Inspired by the concept of entropy, a higher $\alpha_j$ indicates that the segment contains complex, unpredictable patterns (high information density) requiring preservation, whereas a lower $\alpha_j$ suggests a predictable or redundant trend.

Based on these scores, we determine the sampling resolution for each segment. We define the dynamic pooling kernel $k_j$, which determines the compression rate (a larger kernel implies higher compression). The kernel size is computed by balancing the temporal distance bias with the semantic importance bias derived from $\alpha_j$:
\begin{equation}
\label{eq:kernel}
k_j = k_{base} \cdot \underbrace{\left(1 + \frac{j}{n}\right)}_{\text{Distance Decay}} \cdot \underbrace{(1 - \beta \cdot \alpha_j)}_{\text{Semantic Modulation}}
\end{equation}
where $k_{base}$ is the base kernel size, $j$ represents the temporal index (assuming larger $j$ indicates older data), and $\beta \in [0, 1]$ controls the sensitivity to the semantic score. 

The rationale behind Eq. (\ref{eq:kernel}) lies in the dynamic competition between temporal bias and semantic value. The term $(1 + j/n)$ introduces a distance decay effect, imposing a linear penalty on older segments to naturally increase the kernel size and compress distant history. However, this decay is actively counterbalanced by the semantic modulation term $(1 - \beta \cdot \alpha_j)$. 

To provide an intuitive example: consider an older historical segment (large $j$). Under normal circumstances, the temporal penalty would force a large pooling kernel $k_j$, heavily compressing this segment. However, if the importance scorer identifies a critical anomaly or high-entropy pattern within this segment ($\alpha_j \to 1$), the semantic modulation term shrinks significantly. This dynamically overrides the distance penalty, forcing the model to allocate a smaller kernel (dense sampling) to preserve its fine-grained details, regardless of how far back in time it occurred.

Finally, adaptive pooling is applied to compress segment $s_j$ into a sparse representation $s_j^\phi \in \mathbb{R}^{\lambda \times C}$ using the calculated dynamic kernel:
\begin{equation}
s_j^\phi = \text{AdaptivePool}(s_j, \text{kernel}=k_j)
\end{equation}

The compressed segments are then concatenated to form the final sparse sequence $X_i' \in \mathbb{R}^{T \times C}$, which retains critical information efficiently.

\subsection{Cross-Scale Interaction Fusion}
Independent predictions at different scales often ignore the contextual correlation between global trends and local variations. To address this, we replace simple summation with a Cross-Scale Interaction Mixer.

Let $\mathbb{Y}' = \{Y_1', \ldots, Y_N'\}$ be the raw predictions from different scales. We utilize a Gating Network to dynamically weigh the contribution of each scale based on the current context. The fusion weights $W \in \mathbb{R}^{N \times 1}$ are computed as:
\begin{equation}
W = \text{Softmax}(\text{Linear}(\text{Concat}(Y_1', \ldots, Y_N')))
\end{equation}
This allows the model to autonomously decide which scale is most reliable for the current prediction step. The final output is the gated mixture of experts:
\begin{equation}
Y = \sum_{i=1}^N w_i \odot Y_i'
\end{equation}
where $\odot$ denotes element-wise multiplication. This interaction ensures that the final prediction benefits from both the stability of long-term trends and the sensitivity of short-term details.
\section{Theoretical Complexity and Efficiency Analysis}
\label{sec:efficiency}
To address potential concerns regarding the computational overhead introduced by deploying multiple base predictors in our multi-branch framework, we provide a theoretical complexity analysis.

Let $L$ denote the extra-long lookback window, $O$ be the prediction horizon, and $C$ be the number of channels. For a standard Linear-based forecasting model, the temporal linear layer maps the input sequence directly to the prediction, resulting in a spatial parameter complexity and computational cost (FLOPs) of $\mathcal{O}(L \times O)$ per channel.

In the DySCo framework, although the Hierarchical Frequency-Enhanced Decomposition (HFED) introduces $N$ parallel branches (widening the model), the Entropy-Guided Dynamic Sampling (EGDS) aggressively compresses the input sequence from length $L$ to a sparse representation of length $T$, where $T \ll L$. Consequently, the parameter complexity for the $N$ base predictors transforms to $\mathcal{O}(N \times T \times O)$.

As a concrete example, consider an extra-long lookback regime where $L=2440$ and the prediction horizon is $O=336$. The basic Linear model requires $2440 \times 336 = 819,840$ parameters per channel to construct the temporal mapping. In contrast, assuming DySCo utilizes $N=3$ scales and compresses the input to $T=336$, the predictors require only $3 \times 336 \times 336 = 338,688$ parameters per channel. This represents a substantial parameter reduction of approximately $58.7\%$.

While the HFED low-pass filters and the lightweight MLP in the CSIM gating network introduce minor computational operations, they are largely $\mathcal{O}(C)$ or operating on heavily pooled features, which are negligible compared to the temporal matrix multiplications. By ensuring that the compression constraint $N \times T < L$ is satisfied, DySCo effectively neutralizes the $\mathcal{O}(N)$ overhead typically associated with multi-scale ensembles. This theoretical reduction perfectly aligns with our empirical parameter efficiency (as shown in Fig. 7), proving that DySCo achieves superior long-term dependency capture with sub-linear computational scaling relative to the original extra-long context.

Furthermore, the computational advantages of DySCo become exponential when applied to Transformer-based architectures (e.g., PatchTST, iTransformer). Standard self-attention suffers from a quadratic complexity of $\mathcal{O}(L^2 \cdot d)$. By replacing the monolithic long sequence with $N$ sparse representations of length $T$, DySCo curtails this complexity to $\mathcal{O}(N \cdot T^2 \cdot d)$. Using our previous $L=2440$ and $T=336$ setting, this results in an extraordinary 94.3\% reduction in attention-related computations, definitively resolving the memory bottleneck typical of long-context forecasting.


\section{Experiments}
\label{sec:exp}

\subsection{Experimental Setup}
\textbf{Datasets:} We utilized 7 representative time series datasets across 3 domains (transportation, energy, climate). Details are shown in Table \ref{tab:data}.

\begin{table}[htbp]
\centering
\caption{Dataset detailed descriptions. The dataset size is organized as (Train, Validation, Test).}
\label{tab:data}
\resizebox{\linewidth}{!}{
\begin{tabular}{lccc}
\hline
Dataset     & Dim  & Dataset Size            & Information \\ \hline
ETTm1\cite{ett}       & 7    & (34465, 11521, 11521) & Temperature \\
ETTm2\cite{ett}       & 7    & (34465, 11521, 11521) & Temperature \\
ETTh1\cite{ett}       & 7    & (8545, 2881, 2881)    & Temperature \\
ETTh2\cite{ett}       & 7    & (8545, 2881, 2881)    & Temperature \\
Electricity\cite{et} & 321  & (18317, 2633, 5261)   & Electricity\cite{et} \\
Traffic     & 862  & (12185, 1757, 3509)   & Transportation \\
Weather\cite{12}     & 21   & (36792, 5271, 10540)  & Weather      \\ \hline
\end{tabular}
}
\end{table}

\textbf{Baselines \& Competitors:} 
To evaluate the versatility of DySCo across diverse architectures, we conduct experiments on four representative baselines, encompassing both Transformer-based and linear-based models: PatchTST~\cite{11}, iTransformer~\cite{14}, TimeMixer~\cite{15}, and Linear~\cite{16}.

\textbf{Implementation Details \& Hyperparameters:} All experiments were conducted on a single NVIDIA 2080Ti GPU. To ensure full reproducibility, we explicitly detail the key hyperparameter configurations of the DySCo module. For the Hierarchical Frequency-Enhanced Decomposition (HFED), we set the number of scales to $N=3$, with hierarchical lookback horizons configured as $H_i \in \{720, 1440, 2440\}$. The corresponding kernel sizes for the 1D average pooling (low-pass filters) are set inversely to the bandwidth $\sigma_i$, specifically $k_i \in \{8, 24, 36\}$. For the Entropy-Guided Dynamic Sampling (EGDS), the filtered sequences are divided into $n=24$ segments. The base pooling kernel size is set to $k_{base}=8$, and the semantic sensitivity weight is $\beta=0.8$. The overall sampling rate across the module is strictly maintained at $20\%$.

Furthermore, to ensure statistical stability and mitigate the randomness of the learnable sampling and gating components, all reported results of DySCo are averaged over 3 independent runs with different random seeds. 

\textbf{Evaluation Protocol \& Fairness:} A major challenge in evaluating long-context TSF is that basic models typically suffer severe performance degradation when exposed to extra-long sequences due to noise accumulation. To ensure a strictly rigorous and fair comparison, we adopt a conservative evaluation protocol. We fix the lookback window of all DySCo-integrated models at $L=2440$. As empirically validated in our subsequent parameter analysis, this length acts as the ``sweet spot'' where DySCo optimally captures global dependencies without overfitting. In contrast, forcing basic models to blindly operate at $L=2440$ would severely handicap them. Therefore, we conduct an extensive hyperparameter search for all non-DySCo baselines across varying lengths $L \in \{96, 192, 336, 720, 1440, 2440\}$. In our comparative analysis, we report both their performance at length 2440 and their absolute best-tuned results. Demonstrating that DySCo (fixed at 2440) consistently outperforms the optimally tuned configurations of the basic models provides robust evidence of its superiority.

\subsection{Main Results and Comparative Analysis}

Table \ref{tab:main_results} presents the comparison between DySCo-integrated models and their vanilla versions. DySCo consistently boosts performance. For instance, DySCo-integrated TimeMixer achieved an MSE of 0.141 compared to 0.201 for the vanilla version on the Electricity dataset, demonstrating a significant reduction in prediction error. Notably, DySCo with a 2440 lookback window outperforms basic models even when they are tuned to their optimal lookback windows.

\begin{table*}[t]
\centering
\caption{Results of Comparative Experiments. ``Basic(2440)" denotes the baseline model with input length 2440. ``Basic(best)" denotes the baseline with its optimal input length among $\{96, 192, 336, 720,1440,2440\}$. Lower MSE/MAE indicates better accuracy. The best results are highlighted in \textbf{bold}.}
\label{tab:main_results}
\small 
\setlength{\tabcolsep}{4.5pt} 
\renewcommand{\arraystretch}{1.1} 
\begin{tabular}{l | cc cc cc || cc cc cc}
\toprule
\multirow{3}{*}{\textbf{Dataset}} & \multicolumn{6}{c||}{\textbf{TimeMixer}} & \multicolumn{6}{c}{\textbf{PatchTST}} \\ 
\cmidrule(lr){2-7} \cmidrule(lr){8-13}
 & \multicolumn{2}{c}{\textbf{DySCo}} & \multicolumn{2}{c}{Basic(2440)} & \multicolumn{2}{c||}{Basic(best)} & \multicolumn{2}{c}{\textbf{DySCo}} & \multicolumn{2}{c}{Basic(2440)} & \multicolumn{2}{c}{Basic(best)} \\
 & MSE & MAE & MSE & MAE & MSE & MAE & MSE & MAE & MSE & MAE & MSE & MAE \\ 
\midrule
ETT(avg)    & \textbf{0.339} & \textbf{0.369} & 0.445 & 0.450 & 0.357 & 0.389 & \textbf{0.321} & \textbf{0.362} & 0.446 & 0.482 & 0.342 & 0.377 \\
Weather     & \textbf{0.219} & \textbf{0.253} & 0.260 & 0.314 & 0.231 & 0.267 & \textbf{0.221} & \textbf{0.253} & 0.231 & 0.332 & 0.231 & 0.266 \\
Traffic     & \textbf{0.388} & \textbf{0.258} & 0.442 & 0.321 & 0.405 & 0.285 & \textbf{0.400} & \textbf{0.271} & 0.623 & 0.441 & 0.412 & 0.280 \\
Electricity & \textbf{0.141} & \textbf{0.221} & 0.201 & 0.332 & 0.166 & 0.257 & \textbf{0.152} & \textbf{0.231} & 0.189 & 0.272 & 0.162 & 0.255 \\ 
\midrule \midrule
\multirow{3}{*}{\textbf{Dataset}} & \multicolumn{6}{c||}{\textbf{iTransformer}} & \multicolumn{6}{c}{\textbf{Linear}} \\ 
\cmidrule(lr){2-7} \cmidrule(lr){8-13}
 & \multicolumn{2}{c}{\textbf{DySCo}} & \multicolumn{2}{c}{Basic(2440)} & \multicolumn{2}{c||}{Basic(best)} & \multicolumn{2}{c}{\textbf{DySCo}} & \multicolumn{2}{c}{Basic(2440)} & \multicolumn{2}{c}{Basic(best)} \\
 & MSE & MAE & MSE & MAE & MSE & MAE & MSE & MAE & MSE & MAE & MSE & MAE \\ 
\midrule
ETT(avg)    & \textbf{0.357} & \textbf{0.374} & 0.415 & 0.441 & 0.370 & 0.398 & \textbf{0.371} & \textbf{0.413} & 0.404 & 0.428 & 0.391 & 0.421 \\
Weather     & \textbf{0.226} & \textbf{0.265} & 0.277 & 0.318 & 0.239 & 0.274 & \textbf{0.241} & \textbf{0.291} & 0.285 & 0.329 & 0.288 & 0.322 \\
Traffic     & \textbf{0.365} & \textbf{0.264} & 0.371 & 0.271 & 0.399 & 0.282 & \textbf{0.482} & \textbf{0.353} & 0.929 & 0.631 & 0.675 & 0.551 \\
Electricity & \textbf{0.151} & \textbf{0.243} & 0.155 & 0.253 & 0.164 & 0.258 & \textbf{0.185} & \textbf{0.289} & 0.201 & 0.306 & 0.208 & 0.325 \\ 
\bottomrule
\end{tabular}
\end{table*}



\subsection{Mechanism Analysis and Ablation Studies}

\subsubsection{Impact of Sampling Strategies}
A core contribution of DySCo is the Entropy-Guided Dynamic Sampling (EGDS). To validate its necessity, we compared EGDS against other sampling strategies: Uniform Sampling, Random Sampling, and Top-k Frequency Sampling. All the sampling rate is maintained at 20\%. The results in Table \ref{tab:sampling_strategy} show that naive uniform sampling leads to information loss (aliasing), while EGDS effectively retains high-entropy information, resulting in the lowest MSE.


\begin{table}[htbp]
\centering
\caption{Comparison of different sampling strategies across ETT(Avg), Weather, and ECL datasets (Base model: Linear). The best results are highlighted in \textbf{bold}.}
\label{tab:sampling_strategy}
\renewcommand{\arraystretch}{1.1} 
\begin{tabular}{lcccccc} 
\toprule
\multirow{2}{*}{\textbf{Sampling Strategy}} & \multicolumn{2}{c}{ETT (Avg)} & \multicolumn{2}{c}{Weather} & \multicolumn{2}{c}{Electricity} \\ 
\cmidrule(lr){2-3} \cmidrule(lr){4-5} \cmidrule(lr){6-7}
& MSE & MAE & MSE & MAE & MSE & MAE \\ 
\midrule
Uniform Sampling & 0.426 & 0.458 & 0.298 & 0.364 & 0.222 & 0.346 \\
Random Sampling  & 0.589 & 0.612 & 0.347 & 0.401 & 0.258 & 0.397 \\
Top-k Frequency  & 0.412 & 0.498 & 0.283 & 0.347 & 0.221 & 0.342 \\ 
\midrule
\textbf{EGDS} & \textbf{0.371} & \textbf{0.413} & \textbf{0.241} & \textbf{0.291} & \textbf{0.185} & \textbf{0.289} \\ 
\bottomrule
\end{tabular}
\end{table}

\subsubsection{Ablation Study}
We performed a leave-one-out ablation study to quantify the contribution of each component: HFED, EGDS, and CSIM.
\begin{itemize}
    \item \textbf{w/o HFED:} Removed hierarchical decomposition.
    \item \textbf{w/o EGDS:} Remove EGDS without any other sample module.
    \item \textbf{w/o CSIM:} Replaced the gating mixer with linear summation.
\end{itemize}
Table \ref{tab:ablation} demonstrates that removing any component degrades performance, every components contributed to the performance of DySCo.

\begin{table}[htbp]
\centering
\caption{Ablation study of DySCo components on ETT(Avg), Weather, and ECL datasets(Base model: Linear). The best results are highlighted in \textbf{bold}.}
\label{tab:ablation}
\resizebox{0.5\textwidth}{!}{
\begin{tabular}{lcccccc}
\toprule
\multirow{2}{*}{Variant} & \multicolumn{2}{c}{ETT (Avg)} & \multicolumn{2}{c}{Weather} & \multicolumn{2}{c}{Electricity} \\ \cmidrule(r){2-3} \cmidrule(lr){4-5} \cmidrule(l){6-7}
                         & MSE           & MAE           & MSE          & MAE          & MSE            & MAE            \\ \midrule
w/o HFED                 & 0.394         & 0.418         & 0.252        & 0.304        & 0.192          & 0.299          \\
w/o EGDS                 & 0.443         & 0.522         & 0.289        & 0.321        & 0.237          & 0.393          \\
w/o CSIM                 & 0.381         & 0.447         & 0.244        & 0.299        & 0.191          & 0.303          \\ \midrule
\textbf{DySCo (Full)}    & \textbf{0.371} & \textbf{0.413} & \textbf{0.241} & \textbf{0.291} & \textbf{0.185} & \textbf{0.289} \\ \bottomrule
\end{tabular}
}
\end{table}

\begin{figure*}[htbp]
    \centering
    \includegraphics[width=1\linewidth]{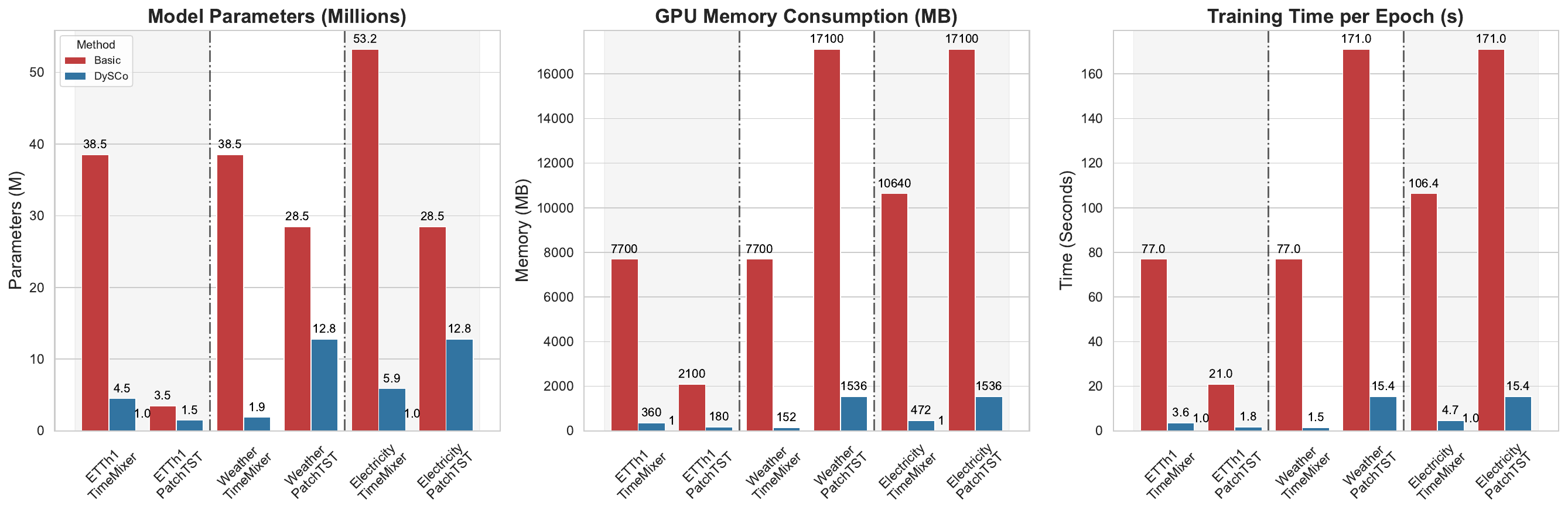}
    \caption{Efficiency comparison of basic models versus DySCo-integrated models across three representative datasets. The evaluation includes model parameters (left), GPU memory consumption (center), and training time per epoch (right). DySCo consistently demonstrates significant reductions across all computational metrics despite utilizing a multi-branch architecture.}
    \label{fig:exp2}
\end{figure*}

\subsection{Visualization of Prediction Results}

To intuitively evaluate the effectiveness of DySCo across various data characteristics, we visualize the prediction results on two representative datasets: ETTh1, which possesses strong long-term evolutionary trends, and Electricity, which is characterized by significant short-term periodicity. The experiments are conducted with a lookback window $L=2440$ and a prediction horizon $O=720$, using the vanilla Linear model as the baseline for comparison.

\begin{figure*}[htbp]
    \centering
    \includegraphics[width=0.85\linewidth]{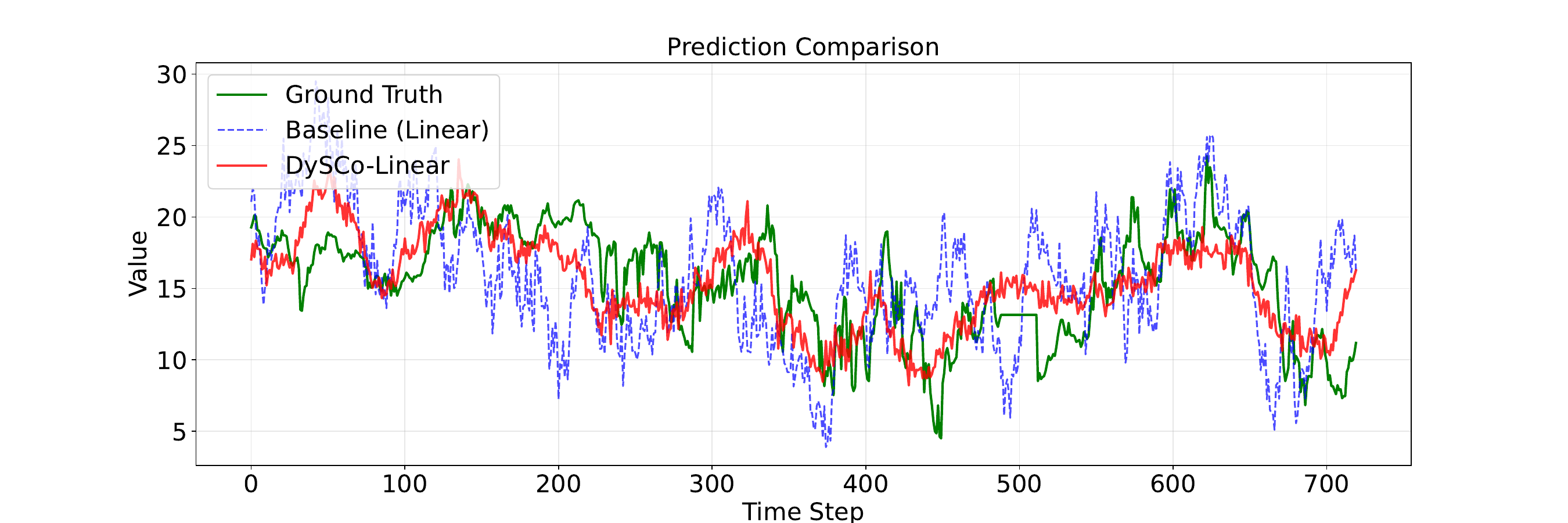}
    \caption{Visualization on ETTh1: DySCo effectively captures the long-term trends underlying the noisy historical data, whereas the baseline model struggles with information redundancy.}
    \label{fig:vis_etth1}
\end{figure*}

\begin{figure*}[htbp]
    \centering
    \includegraphics[width=0.85\linewidth]{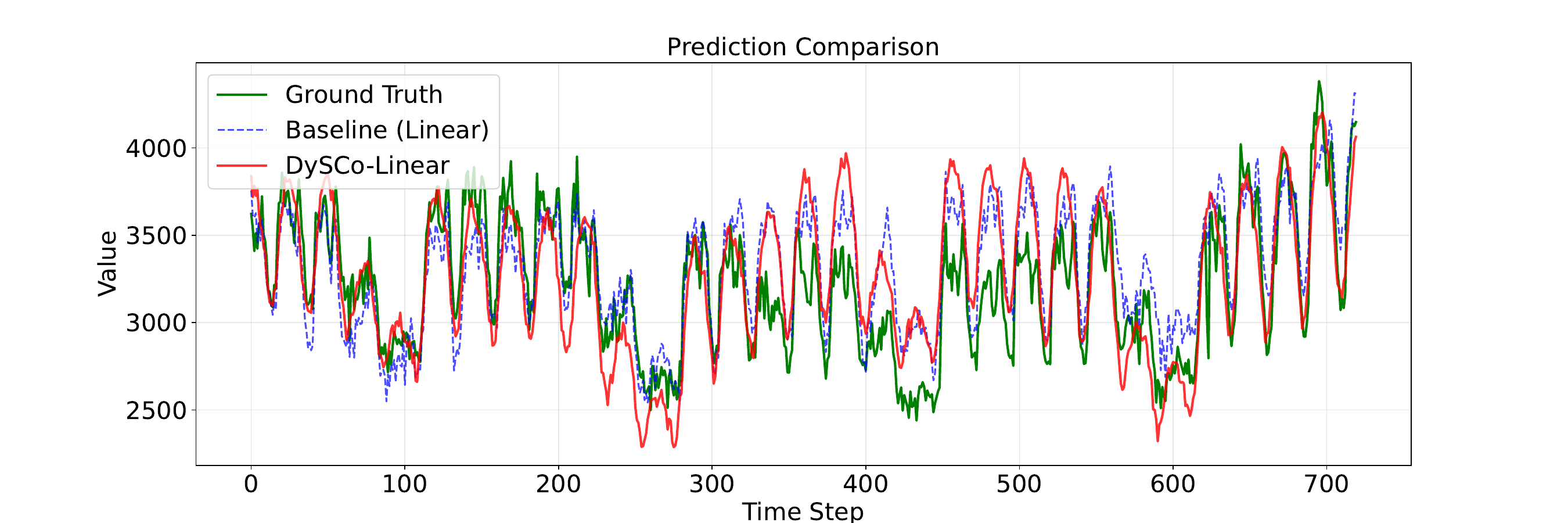}
    \caption{Visualization on Electricity: DySCo maintains high prediction fidelity consistent with the baseline in short-term periodic scenarios, demonstrating its robustness.}
    \label{fig:vis_ecl}
\end{figure*}

\begin{figure}[htbp]
    \centering
    \includegraphics[width=0.9\linewidth]{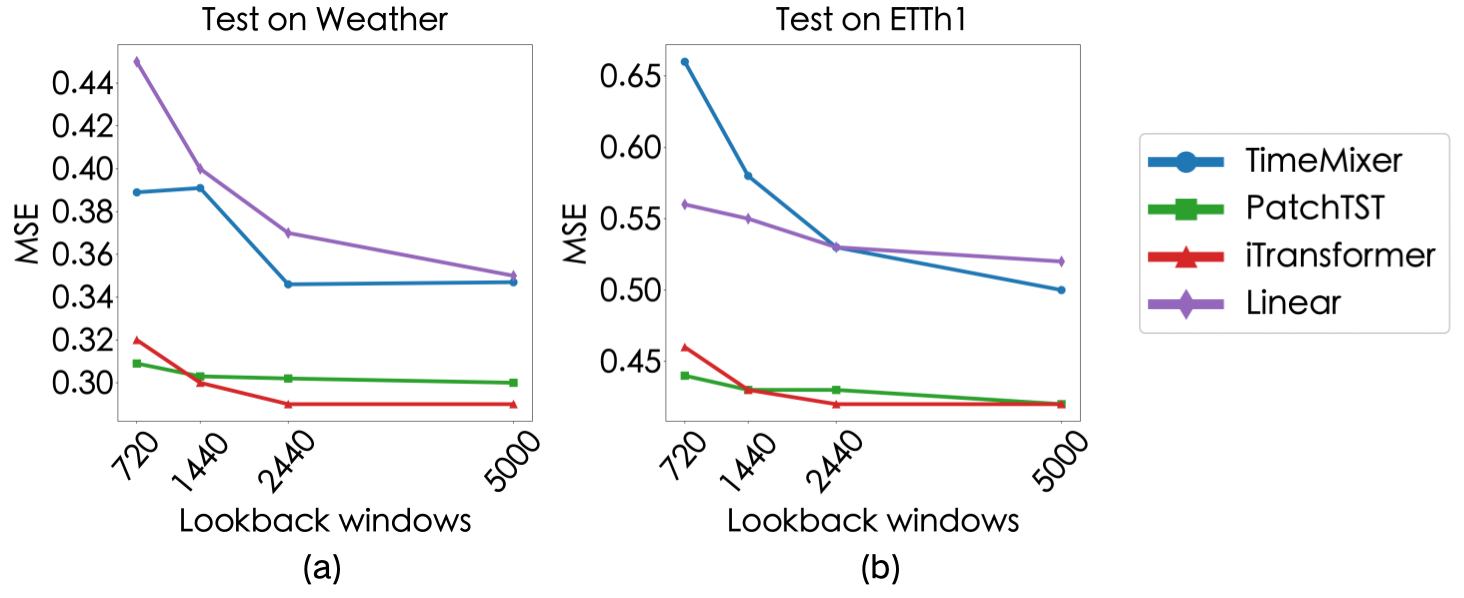}
    \caption{Performance Analysis with Varying Lookback Windows.}
    \label{fig:exp1}
\end{figure}

As illustrated in Fig. \ref{fig:vis_etth1}, DySCo demonstrates a substantial performance gain on the ETTh1 dataset. This improvement is primarily attributed to the EGDS mechanism, which identifies and preserves high-entropy segments that contain critical semantic information for long-term forecasting while filtering out redundant noise. In contrast, the baseline model tends to deviate from the ground truth, as it fails to effectively distinguish global trends from the cluttered information within the extra-long lookback window.

In contrast, on the Electricity dataset (Fig. \ref{fig:vis_ecl}), where short-term patterns dominate, the performance gain from DySCo is relatively marginal. However, considering our previous complexity analysis—which demonstrates that DySCo drastically reduces memory consumption and training time—this marginal accuracy gain is actually a substantial victory. It proves that even in short-term dominated scenarios, DySCo successfully maintains baseline-level accuracy while severely slashing computational overhead. 

\subsection{Parameter Analysis}

\subsubsection{Sensitivity to Lookback Window and Hyperparameters}
We further explored the impact of lookback window lengths (Fig. \ref{fig:exp1}). DySCo models show consistent improvement as the lookback window increases from 720 to 5000.

\subsubsection{Computational Efficiency}
Finally, we evaluated the end-to-end computational efficiency and overhead of the proposed framework. As shown in Figure \ref{fig:exp2}, although DySCo incorporates multiple parallel predictors across different scales, the aggressive down-sampling mechanism introduced by EGDS significantly reduces the overall computational burden. Compared to the basic models operating on the full extra-long lookback window, DySCo achieves a drastic reduction not only in the parameter count but also in GPU memory consumption and training time per epoch. This efficiency gain is particularly pronounced in Transformer-based architectures (e.g., PatchTST), where DySCo effectively mitigates the quadratic memory complexity associated with long-sequence self-attention. These results confirm that DySCo is a highly efficient, lightweight module capable of breaking the performance-efficiency bottleneck in long-context forecasting.

\section{Conclusion}
\label{sec:conclusion}
We propose DySCo, a universal plug-and-play framework that enhances long-term time series forecasting by adaptively distilling critical dependencies from redundant, noisy historical data. By replacing fixed heuristics with a learnable semantic compression paradigm, DySCo enables mainstream models to effectively utilize extra-long lookback windows with significantly reduced computational costs. Extensive experiments across seven datasets and four architectures confirm that DySCo consistently boosts predictive accuracy and parameter efficiency, offering a robust solution for capturing long-range correlations.

\section*{Acknowledgment}
During the preparation of this work, the author used Gemini (Google) to assist with language editing and LaTeX formatting.

\bibliographystyle{IEEEtran}
\bibliography{ref}

@article{mamba_tsf2025,
      title={MambaTS: Improved Selective State Space Models for Long-term Time Series Forecasting}, 
      author={Xiuding Cai and Yaoyao Zhu and Xueyao Wang and Yu Yao},
      year={2024},
      eprint={2405.16440},
      archivePrefix={arXiv},
      primaryClass={cs.LG},
      url={https://arxiv.org/abs/2405.16440}, 
}

@misc{timemixer++,
      title={TimeMixer++: A General Time Series Pattern Machine for Universal Predictive Analysis}, 
      author={Shiyu Wang and Jiawei Li and Xiaoming Shi and Zhou Ye and Baichuan Mo and Wenze Lin and Shengtong Ju and Zhixuan Chu and Ming Jin},
      year={2025},
      eprint={2410.16032},
      archivePrefix={arXiv},
      primaryClass={cs.LG},
      url={https://arxiv.org/abs/2410.16032}, 
}

@inproceedings{micn,
  title={MICN: Multi-scale Local and Global Context Modeling for Long-term Series Forecasting},
  author={Huiqiang Wang and Jian Peng and Feihu Huang and Jince Wang and Junhui Chen and Yifei Xiao},
  booktitle={International Conference on Learning Representations},
  year={2023},
  url={https://api.semanticscholar.org/CorpusID:259298592}
}

@misc{frequency,
      title={Time Series Analysis in Frequency Domain: A Survey of Open Challenges, Opportunities and Benchmarks}, 
      author={Qianru Zhang and Yuting Sun and Honggang Wen and Peng Yang and Xinzhu Li and Ming Li and Kwok-Yan Lam and Siu-Ming Yiu and Hongzhi Yin},
      year={2025},
      eprint={2504.07099},
      archivePrefix={arXiv},
      primaryClass={cs.CE},
      url={https://arxiv.org/abs/2504.07099}, 
}

@inbook{1,
author = {Ao, Xiang},
title = {Research on the Stock Price Prediction Model Based on Large-Scale Transactions},
year = {2025},
isbn = {9798400718748},
publisher = {Association for Computing Machinery},
address = {New York, NY, USA},
url = {https://doi.org/10.1145/3773365.3773427},
booktitle = {Proceedings of the 2025 8th International Conference on Computer Information Science and Artificial Intelligence},
pages = {392–396},
numpages = {5}
}

@article{2,
  title={Interpretable Weather Forecasting For Worldwide Stations with a Unified Deep Model},
  author={Wu, Haixu and Zhou, Hang and Long, Mingsheng and others},
  journal={Nature Machine Intelligence},
  volume={5},
  number={6},
  pages={602--611},
  year={2023},
  publisher={Nature Publishing Group UK London},
  doi={10.1038/s42256-023-00667-9},
  url={https://doi.org/10.1038/s42256-023-00667-9}
}

@article{3,
  title={Estimation of the Arrival Times of Seismic Waves by Multivariate Time Series Model},
  author={Takanami, T. and Kitagawa, G.},
  journal={Annals of the Institute of Statistical Mathematics},
  volume={43},
  number={3},
  pages={407--433},
  year={1991},
  publisher={Springer},
  doi={10.1007/BF00053364},
  url={https://doi.org/10.1007/BF00053364}
}

@INPROCEEDINGS{4,
  author={Yang, Yang and Cao, Longbing},
  booktitle={2023 International Joint Conference on Neural Networks (IJCNN)}, 
  title={MTSNet: Deep Probabilistic Cross-multivariate Time Series Modeling with External Factors for COVID-19}, 
  year={2023},
  volume={},
  number={},
  pages={1-10},
  doi={10.1109/IJCNN54540.2023.10191636}}

@misc{5,
      title={TimesNet: Temporal 2D-Variation Modeling for General Time Series Analysis}, 
      author={Haixu Wu and Tengge Hu and Yong Liu and Hang Zhou and Jianmin Wang and Mingsheng Long},
      year={2023},
      eprint={2210.02186},
      archivePrefix={arXiv},
      primaryClass={cs.LG},
      url={https://arxiv.org/abs/2210.02186}, 
}

@article{6,
  title={Time Series Analysis: Forecasting and Control},
  author={G. E. P. Box and Gwilym M. Jenkins and Gregory C. Reinsel and Greta M. Ljung},
  journal={The Statistician},
  year={1978},
  volume={27},
  pages={265-265},
  url={https://api.semanticscholar.org/CorpusID:62556607}
}

@ARTICLE{7,
  author={Hochreiter, Sepp and Schmidhuber, Jürgen},
  journal={Neural Computation}, 
  title={Long Short-Term Memory}, 
  year={1997},
  volume={9},
  number={8},
  pages={1735-1780},
  doi={10.1162/neco.1997.9.8.1735}}

@inproceedings{8,
  title={On the Properties of Neural Machine Translation: Encoder–Decoder Approaches},
  author={Kyunghyun Cho and Bart van Merrienboer and Dzmitry Bahdanau and Yoshua Bengio},
  booktitle={SSST@EMNLP},
  year={2014},
  url={https://api.semanticscholar.org/CorpusID:11336213}
}

@misc{9,
      title={N-BEATS: Neural Basis Expansion Analysis For Interpretable Time Series Forecasting}, 
      author={Boris N. Oreshkin and Dmitri Carpov and Nicolas Chapados and Yoshua Bengio},
      year={2020},
      eprint={1905.10437},
      archivePrefix={arXiv},
      primaryClass={cs.LG},
      url={https://arxiv.org/abs/1905.10437}, 
}

@article{10,
title = {DeepAR: Probabilistic Forecasting with Autoregressive Recurrent Networks},
journal = {International Journal of Forecasting},
volume = {36},
number = {3},
pages = {1181-1191},
year = {2020},
issn = {0169-2070},
doi = {https://doi.org/10.1016/j.ijforecast.2019.07.001},
url = {https://www.sciencedirect.com/science/article/pii/S0169207019301888},
author = {David Salinas and Valentin Flunkert and Jan Gasthaus and Tim Januschowski}
}

@misc{11,
      title={A Time Series is Worth 64 Words: Long-term Forecasting with Transformers}, 
      author={Yuqi Nie and Nam H. Nguyen and Phanwadee Sinthong and Jayant Kalagnanam},
      year={2023},
      eprint={2211.14730},
      archivePrefix={arXiv},
      primaryClass={cs.LG},
      url={https://arxiv.org/abs/2211.14730}, 
}

@inproceedings{12,
      title={Autoformer: Decomposition Transformers with Auto-Correlation for Long-Term Series Forecasting}, 
      author={Haixu Wu and Jiehui Xu and Jianmin Wang and Mingsheng Long},
      year={2022},
      eprint={2106.13008},
      archivePrefix={arXiv},
      primaryClass={cs.LG},
      url={https://arxiv.org/abs/2106.13008}, 
}

@misc{14,
      title={iTransformer: Inverted Transformers Are Effective for Time Series Forecasting}, 
      author={Yong Liu and Tengge Hu and Haoran Zhang and Haixu Wu and Shiyu Wang and Lintao Ma and Mingsheng Long},
      year={2024},
      eprint={2310.06625},
      archivePrefix={arXiv},
      primaryClass={cs.LG},
      url={https://arxiv.org/abs/2310.06625}, 
}

@article{15,
  title={TimeMixer: Decomposable Multiscale Mixing for Time Series Forecasting},
  author={Shiyu Wang and Haixu Wu and Xiao Long Shi and Tengge Hu and Huakun Luo and Lintao Ma and James Y. Zhang and Jun Zhou},
  journal={ArXiv},
  year={2024},
  volume={abs/2405.14616},
  url={https://api.semanticscholar.org/CorpusID:269982658}
}

@misc{16,
      title={Are Transformers Effective for Time Series Forecasting?}, 
      author={Ailing Zeng and Muxi Chen and Lei Zhang and Qiang Xu},
      year={2022},
      eprint={2205.13504},
      archivePrefix={arXiv},
      primaryClass={cs.AI},
      url={https://arxiv.org/abs/2205.13504}, 
}

@INPROCEEDINGS{xlinear,
  author={Ao, Xiang},
  booktitle={2025 5th International Conference on Artificial Intelligence, Automation and High Performance Computing (AIAHPC)}, 
  title={XLinear: Frequency-Enhanced MLP with CrossFilter for Robust Long-Range Forecasting}, 
  year={2025},
  volume={},
  number={},
  pages={426-433},
  doi={10.1109/AIAHPC66801.2025.11290064}}

@ARTICLE{m1,
  author={Vetterli, M. and Marziliano, P. and Blu, T.},
  journal={IEEE Transactions on Signal Processing}, 
  title={Sampling Signals with Finite Rate of Innovation}, 
  year={2002},
  volume={50},
  number={6},
  pages={1417-1428},
  doi={10.1109/TSP.2002.1003065}}

@article{m2,
  title={Influence of Different Data Interpolation Methods for Sparse Data on the Construction Accuracy of Electric Bus Driving Cycle},
  author={Wang, Xingxing and Ye, Peilin and Deng, Yelin and Yuan, Yinnan and Zhu, Yu and Ni, Hongjun},
  journal={Electronics},
  volume={12},
  number={6},
  pages={1377},
  year={2023},
  publisher={MDPI},
  doi={10.3390/electronics12061377},
  url={https://doi.org/10.3390/electronics12061377}
}

@misc{ett,
      title={Informer: Beyond Efficient Transformer for Long Sequence Time-Series Forecasting}, 
      author={Haoyi Zhou and Shanghang Zhang and Jieqi Peng and Shuai Zhang and Jianxin Li and Hui Xiong and Wancai Zhang},
      year={2021},
      eprint={2012.07436},
      archivePrefix={arXiv},
      primaryClass={cs.LG},
      url={https://arxiv.org/abs/2012.07436}, 
}

@article{et,
  title={Modeling Long- and Short-Term Temporal Patterns with Deep Neural Networks},
  author={Guokun Lai and Wei-Cheng Chang and Yiming Yang and Hanxiao Liu},
  journal={The 41st International ACM SIGIR Conference on Research \& Development in Information Retrieval},
  year={2017},
  url={https://api.semanticscholar.org/CorpusID:4922476}
}

\end{document}